\pdfoutput=1

\documentclass[11pt]{article}

\usepackage[final]{acl}

\usepackage{times}
\usepackage{latexsym}

\usepackage{amsmath}
\usepackage{algorithm}
\usepackage{algpseudocode}
\usepackage{listings}

\usepackage[T1]{fontenc}

\usepackage[utf8]{inputenc}

\usepackage{microtype}

\usepackage{inconsolata}

\usepackage{graphicx}

\title{MedOrchestra: A Hybrid Cloud-Local LLM Approach for Clinical Data Interpretation}

\author{
 \textbf{Sihyeon Lee\textsuperscript{1}},
 \textbf{Hyunjoo Song\textsuperscript{2}},
 \textbf{Jong-chan Lee\textsuperscript{3}},
 \textbf{Yoon Jin Lee\textsuperscript{3}},
 \textbf{Boram Lee\textsuperscript{3}},
\\
 \textbf{Hee-Eon Lim\textsuperscript{3}},
 \textbf{Dongyeong Kim\textsuperscript{3}},
 \textbf{Jinwook Seo\textsuperscript{1}},
 \textbf{Bohyoung Kim\textsuperscript{4}}
\\
\\
 \textsuperscript{1}Seoul National University,
 \textsuperscript{2}Soongsil University,
 \textsuperscript{3}Seoul National Univeristy Bundang Hospital,
\\
 \textsuperscript{4}Hankuk University of Foreign Studies
\\
}

\begin{document}
\maketitle
\begin{abstract}
Deploying large language models (LLMs) in clinical settings faces critical trade-offs: cloud LLMs, with their extensive parameters and superior performance, pose risks to sensitive clinical data privacy, while local LLMs preserve privacy but often fail at complex clinical interpretation tasks. We propose MedOrchestra, a hybrid framework where a cloud LLM decomposes complex clinical tasks into manageable subtasks and prompt generation, while a local LLM executes these subtasks in a privacy-preserving manner. Without accessing clinical data, the cloud LLM generates and validates subtask prompts using clinical guidelines and synthetic test cases. The local LLM executes subtasks locally and synthesizes outputs generated by the cloud LLM. We evaluate MedOrchestra on pancreatic cancer staging using 100 radiology reports under NCCN guidelines. On free-text reports, MedOrchestra achieves 70.21\% accuracy, outperforming local model baselines (without guideline: 48.94\%, with guideline: 56.59\%) and board-certified clinicians (gastroenterologists: 59.57\%, surgeons: 65.96\%, radiologists: 55.32\%). On structured reports, MedOrchestra reaches 85.42\% accuracy, showing clear superiority across all settings.
\end{abstract}

\section{Introduction}
\begin{figure*}[t]
    \includegraphics[width=1.0\textwidth]{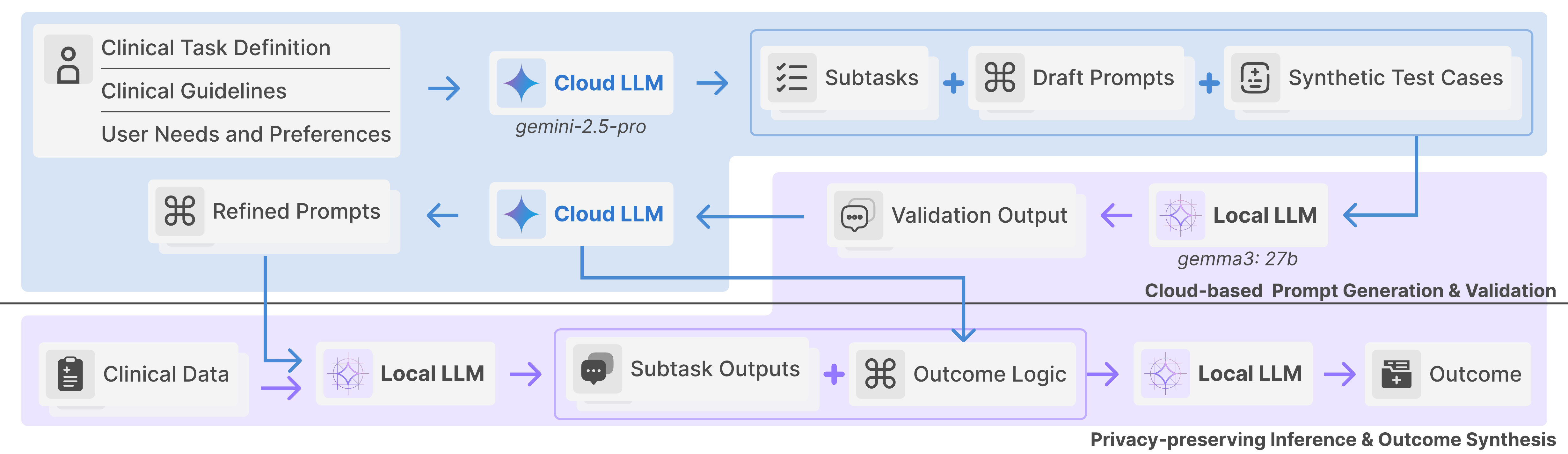} \hfill
    \caption {\textbf{Overview of the MedOrchestra framework.} The system operates in two phases: (1) cloud-based prompt generation \& validation, where a cloud LLM decomposes the user-defined clinical task into subtasks, generates draft prompts, and validates them using synthetic test cases; and (2) privacy-preserving inference \& outcome synthesis, where a local LLM applies the refined prompts to real clinical data to extract subtask outputs, which are then synthesized into a final outcome.}
    \label{fig:overview}
\end{figure*}

Free-text clinical reports, particularly those produced in radiology and pathology, play a central role in clinical decision-making. These unstructured reports contain rich and complex clinical information that supports patient diagnosis, cancer staging, treatment planning, and overall care management \citep{raghavan2014unstructured}. Furthermore, the composition of these reports is often influenced by established clinical protocols and standardized guidelines, which help ensure consistency and medical accuracy.

While free-text clinical reports contain vast amounts of valuable clinical information, their unstructured language patterns and diverse expressions often make it challenging to quickly identify or extract the necessary information in actual clinical settings \citep{sedlakova2023challenges}. This limitation can hinder clinical efficiency and consistency, which has led to the growing adoption of Natural Language Processing (NLP) technologies as a complementary solution.

Conventional NLP methods, including rule-based systems and various machine learning algorithms such as SVM, CRF, and Random Forest, have been applied to extract clinical information from free-text radiology reports \citep{nobel2024nlp, kumbhakarna2020nlp}. However, their performance remains limited by institutional differences in documentation styles and challenges in handling uncertainty and implicit language, suggesting the need for more context-aware approaches. In addition, these methods often require task and data-specific training and manual feature engineering, which limits their scalability and adaptability across different clinical use cases.

In light of these limitations, recent advances in large language models (LLMs) have drawn attention for their ability to overcome many of the challenges faced by conventional NLP methods. Unlike earlier approaches, LLMs are pretrained on massive text corpora and demonstrate strong capabilities in understanding context, handling uncertainty, and generalizing across diverse clinical tasks with minimal task-specific adaptation \citep{manathunga2023aligninglargelanguagemodels, yang2025exploringlargelanguagemodels}. These strengths make them particularly well-suited for processing complex and variable free-text radiology reports, especially when aligned with established clinical guidelines.

Many state-of-the-art cloud LLMs (e.g., GPT-4o \citep{openai2024gpt4o}, Gemini 2.5 Pro \citep{deepmind2025gemini25}) available through commercial cloud platforms are characterized by extremely large parameter sizes and extended context windows. These features allow them to process complex clinical narratives more effectively. Several studies have shown strong performance in tasks such as extracting decision-critical information, structuring free-text reports, and supporting evidence-based clinical reasoning \citep{reichenpfader2023llm, vrdoljak2025review, wu2024guiding}. Despite these strengths, cloud LLMs are rarely used in real-world clinical settings. The main reason is patient privacy. Sending sensitive clinical data to external servers is often restricted by institutional polices and legal regulations \citep{marks2023ai}.

To address privacy concerns, research has emerged exploring the use of local LLMs (e.g., Llama \citep{meta2024llama4}, Gemma \citep{google2025gemma3}) in clinical environments \citep{vaid2024local, wiest2024privacy}. While some of these local LLMs have large parameter counts, their practical deployment in clinical settings is often limited by hardware constraints and high implementation costs. Consequently, smaller models are typically employed, which may result in performance degradation in complex clinical tasks that require sophisticated contextual understanding and precision \citep{wang2024comprehensivesurveysmalllanguage}.

To address such performance degradation, researchers have explored various techniques, including fine-tuning \citep{hou2025fine}, retrieval-augmented generation (RAG) \citep{ke2025retrieval}, and various prompt engineering strategies \citep{maharjan2024openmedlm}. However, the application of these methods in clinical settings remains limited. Obtaining high-quality clinical data and annotations is challenging, and even minor changes often require re-running the entire process, making these approaches burdensome and difficult to apply in real-world clinical settings \citep{dennstadt2025implementing}.

In response to the inherent limitations of cloud and local LLMs, we propose MedOrchestra, a hybrid cloud-local LLM framework. MedOrchestra combines the strengths of both cloud and local LLMs. The cloud LLM handles complex language tasks that require high performance and long-context processing, and the local LLM ensures privacy-preserving inference by keeping sensitive clinical data on-site. This hybrid structure allows tasks to be divided based on data sensitivity and computational needs. An overview of the MedOrchestra framework is shown in Figure~\ref{fig:overview}.

In MedOrchestra, the cloud LLM acts as a meta-orchestrator. Upon receiving the clinical task definition, relevant guidelines, and user needs, the cloud LLM decomposes the overall task into a set of manageable subtasks that can be handled by the local LLM. It then generates corresponding draft prompts for each subtask, along with synthetic test cases to support prompt validation. Furthermore, the cloud LLM defines the outcome logic, the rules for aggregating the outputs from individual subtasks to produce the final clinical outcome. This process leverages the cloud LLM's strong performance and contextual reasoning capabilities while avoiding exposure of any sensitive clinical data at this stage.

The local LLM in MedOrchestra serves as the primary inference engine for handling sensitive clinical data. It begins by using the subtasks, draft prompts, and synthetic test cases generated by the cloud LLM to produce a validation output, which includes predicted answers and reasoning traces. This output is sent back to the cloud LLM, which compares the results against expected outcomes and, if necessary, refines the prompts to produce an improved version. Once validation is complete, the local LLM uses the refined prompts to make inferences on actual clinical data. Each subtask generates output and then applies the outcome logic, originally defined by the cloud LLM, to integrate the subtask results and derive the final clinical outcome.

To evaluate its applicability in real clinical settings, MedOrchestra was applied to clinical staging tasks using 100 radiology reports (50 free-text and 50 structured format) from pancreatic cancer patients based on the NCCN clinical guideline \footnote{\url{https://www.nccn.org/guidelines/guidelines-detail?category=1&id=1455}}. Performance was compared against a local LLM baseline (with and without clinical guidelines) as well as three board-certified gastroenterology, surgery, and radiology specialists. MedOrchestra achieved superior accuracy across all comparisons, demonstrating its suitability for clinical guideline-based interpretation of free-text reports while protecting sensitive clinical data.

\section{Related Work}

\subsection{LLMs for Guideline-Driven Interpretation of Radiology Reports}
Recent efforts have actively explored the use of LLMs to interpret clinical free-text, such as radiology reports, according to clinical guidelines. For example, studies based on models like GPT-4, Med-PaLM, and Llama have demonstrated the utility of LLMs in tasks such as staging estimation from radiology reports, summarizing key findings, and structuring lesion information \citep{gu2024gpt4, zhou2024largemodeldrivenradiology, hartsock2025improving}. Notably, recent research has introduced prompt design strategies and evaluation methods that incorporate standardized clinical guidelines such as NCCN or BI-RADS into model responses \citep{kim2025conversion, coBI2024}.
However, most approaches rely on single LLM systems, and when using cloud LLMs, sensitive clinical data must be transmitted externally, making it difficult to ensure privacy. Conversely, when using local LLMs, additional methods such as fine-tuning \citep{chen2024finetuninginhouselargelanguage} or RAG \citep{arasteh2024radioragfactuallargelanguage} are required, resulting in task- or data-specific approaches that are difficult to deploy in real-world environments.

\subsection{Planner–Executor Orchestration with LLMs}
Several works in general NLP have proposed orchestration frameworks in which a planner LLM decomposes tasks and delegates subtasks to smaller models or external tools \citep{schick2023toolformerlanguagemodelsteach, khot2023decomposedpromptingmodularapproach}. This architecture improves modularity and supports data protection by separating sensitive data from the planner, which is especially important in clinical NLP governed by regulations like HIPAA and GDPR.

However, adoption in clinical NLP remains limited due to technical challenges in data separation, lack of annotated datasets, and the complexity of integrating domain-specific workflows. \citep{suster-etal-2017-short, nam-etal-2019-surf}

In MedOrchestra, we assign guideline-based reasoning and task decomposition to a cloud-based planner, while keeping PHI-sensitive inference within a local executor. This setup balances high performance for complex tasks with patient privacy and real-world deployability.

\section{Method}
\subsection{Overview}
MedOrchestra is a hybrid framework that separates clinical task orchestration from data-sensitive inference. As shown in Fig~\ref{fig:overview}, the system operates in two phases: (1) a cloud-based prompt generation and validation phase, and (2) a local inference and outcome synthesis phase. The following sections detail each phase.

\subsection{Clinical Task Input and Subtask Decomposition}
\begin{equation}
    \label{eq:task_input}
    \mathcal{T} = (\tau, \mathcal{G}, \mathcal{U})
\end{equation}
\begin{equation}
    \label{eq:task_decompose}
    \mathcal{S}, \mathcal{P}_\text{draft}, \mathcal{L} = \mathrm{CloudLLM}(\mathcal{T})
\end{equation}
We begin by formalizing the input to the MedOrchestra framework as a triplet \(\mathcal{T} \), consisting of three components: the clinical task description \(\tau \), the associated clinical guideline \(\mathcal{G}\), and a set of user-defined preferences \(\mathcal{U}\). This is represented in Equation~\ref{eq:task_input}.

Here, \(\tau \) typically defines the high-level reasoning goal (e.g., determine clinical staging), \(\mathcal{G}\) denotes the clinical guideline document (e.g., NCCN, AJCC), and \(\mathcal{U} \) encodes user-defined preferences such as desired output format, subtask granularity, or inclusion/exclusion of specific entity types.

Based on this input, the cloud LLM generates three key outputs: a set of subtasks \(\mathcal{S}\), corresponding draft prompts \(\mathcal{P}_\text{draft}\), and a rule-based synthesis logic \(\mathcal{L}\) that defines how subtask outputs are combined into final task outcomes. This process is summarized in Equation~\ref{eq:task_decompose}.

Each subtask \(s_i \in \mathcal{S}\) represents an independent unit of clinical reasoning required to complete the overall task. These subtasks are not predefined but are instead inferred by the cloud LLM based on the full task input Equation~\ref{eq:task_input}. This decomposition allows the system to isolate modular reasoning components, such as primary tumor location, detecting metastatic spread, or evaluating vessel involvement, that can be executed independently by a local LLM.

Once the set of subtasks \(\mathcal{S}\) is established, the cloud LLM constructs a corresponding draft system prompt \(p_i^\text{draft} \in \mathcal{P}_\text{draft}\) for each subtask \(s_i\). These prompts are generated under the assumption that the local LLM lacks access to the \(\tau, \mathcal{G}, \) or any global context. As such, each prompt \(p_i^\text{draft}\) must be self-contained: it includes a natural language task description, relevant background derived from \(\mathcal{G}\), and formatting instructions aligned with \(\mathcal{U}\). This design ensures that each prompt can be executed independently in a restricted local environment.

\subsection{Prompt Validation with Synthetic Test Cases}
To ensure that each draft prompt is interpretable and executable by the local LLM, MedOrchestra performs prompt validation using synthetic test cases. These synthetic inputs are generated by the cloud LLM without any access to real clinical data. Instead, they are constructed by instantiating clinically plausible scenarios directly from the guideline \(\mathcal{G}\) and \(\tau\), yielding inputs that reflect key decision points while preserving data privacy.
Formally, for each subtask \(s_i\), the cloud LLM generates a set of synthetic examples \(\mathcal{X}_{syn}^{(i)}\) and corresponding expected outputs \(\mathcal{Y}_{syn}^{(i)}\) as:
\begin{equation}
    \label{eq:synthetic_input}
    \mathcal{X}_{syn}^{(i)}, \mathcal{Y}_{syn}^{(i)} = GenerateSynthetic(s_i,\mathcal{G})
\end{equation}
Each synthetic input \(x_{syn}^{(i)} \in \mathcal{X}_{syn}^{(i)}\) is then paired with a draft prompt \(p_i^\text{draft}\), and passed to the local LLM for evaluation. The model is expected to generate output \(y\) and reasoning \(r\):
\begin{equation}
    \label{eq:synthetic_evaluation}
    (r_\text{val}^{(i)}, y_\text{val}^{(i)}) = \mathrm{LocalLLM}(p_i^\text{draft}, x_\text{syn}^{(i)})
\end{equation}

A prompt is considered valid only if the predicted output \(y_{val}^{(i)}\) aligns with the expected values defined in \(\mathcal{Y}_{syn}^{(i)}\). This validation process ensures not only correctness but also interpretability, making it easier to detect ambiguous instructions or faulty reasoning induced by the prompt.

\subsection{Prompt Refinement}

\begin{algorithm}[H]
\caption{prompt refinement loop}
\begin{algorithmic}[1]
\For{each subtask $s_i$}
    \State $p_i \gets p_i^\text{draft}$
    \While{validation accuracy on $\text{TestSet}_i$ $<$ 80\%}
        \State $(r_\text{val}, y_\text{val}) \gets \text{LocalLLM}(p_i, x_\text{syn})$
        \State $p_i \gets \text{RefinePrompt}(p_i, r_\text{val})$
    \EndWhile
    \State $p_i^\text{refined} \gets p_i$
\EndFor
\end{algorithmic}
\end{algorithm}

If the predicted output \(y_{val}^{(i)}\) does not match the expected value \(y_{syn}^{(i)}\), the corresponding reasoning trace \(r_{val}^{(i)}\) is reviewed to identify potential causes of failure, such as ambiguous task phrasing, incomplete guideline context, or formatting issues. Based on this analysis, the cloud LLM refines the draft prompt \(p_i^\text{draft}\), yielding an updated version \(p_i^\text{refined}\) that better guides the local model toward the intended behavior. The revised prompt is then re-evaluated on the same synthetic test set. This refinement loop continues until the prompt consistently passes 80\% of the test cases.

\subsection{Inference on Clinical Data and Outcome Synthesis}
Once the refined prompts \(p_i^\text{refined}\) for all subtasks are finalized, the system proceeds to perform inference on real clinical data. For each patient document \(d^{(j)}\), the local LLM executes each subtasks \(s_i \in \mathcal{S}\) independently using the corresponding refined prompt:
\begin{equation}
    \label{eq:refined_inference}
    f_i^{(j)} = LocalLLM(p_i^{refined}, d^{(j)})
\end{equation}
This process yields a set of subtask-specific outputs:
\begin{equation}
    \label{eq:subtask_outputs}
    \mathcal{F}^{(j)} = \{f_1^{(j)}, f_2^{(j)}, ..., f_n^{(j)}\}
\end{equation}
where each \(f_i^{(j)}\) represents a discrete clinical feature or intermediate decision. Once all subtask outputs are collected, the system applies the synthesis logic \(\mathcal{L}\), previously generated by the cloud LLM, to derive the final task outcome:
\begin{equation}
    \label{eq:task_outcome}
    y^{(j)} = Synthesize(\mathcal{F}^{(j)}, \mathcal{L})
\end{equation}
Equation~\ref{eq:task_outcome} formalizes how the subtask outputs \(\mathcal{F}^{(j)}\) are synthesized into a final task outcome using the logic \(\mathcal{L}\), which is derived from the clinical guideline \(\mathcal{G}\). The logic encodes how combinations of intermediate features, such as abnormal findings or clinically significant conditions, inform the final decision.

To account for potential variability in local LLM outputs, the inference process is repeated \(T\) times for each clinical document \(d^{(j)}\), resulting in a set of candidate outcomes:

\[
\mathcal{Y}^{(j)} = \{y_1^{(j)}, y_2^{(j)}, \dots, y_T^{(j)}\}
\]

The final prediction \(\hat{y}^{(j)}\) is selected by majority voting over \(\mathcal{Y}^{(j)}\):

\[
\hat{y}^{(j)} = \mathrm{MajorityVote}(\mathcal{Y}^{(j)})
\]

This strategy enhances the robustness of the final outcome by mitigating the effects of stochastic generation and occasional reasoning errors during local inference.

\section{Experiments}
\subsection{Dataset and Annotation}

We constructed a clinical staging dataset using 100 abdominal imaging reports from patients diagnosed with pancreatic cancer at a tertiary teaching hospital in South Korea between 2003 and 2018. The dataset includes CT and MRI reports, and we fully de-identified all data following institutional guidelines. The hospital's Institutional Review Board (IRB) approved the study protocol, where the data were collected.

The dataset comprises 50 free-text and 50 structured-form reports, reflecting the diversity of radiological documentation styles in real-world clinical settings. The reports were written in Korean and English, as is common in bilingual clinical documentation practices in Korea. We used only the body of each report for all experiments, excluding the Conclusion section. This design aimed to simulate common clinical workflows, where non-radiologist specialists often make staging decisions based solely on the narrative report without direct image review.

We inferred ground truth (GT) staging labels from the original Conclusion sections written by board-certified radiologists during routine care. While these conclusions did not explicitly assign one of the NCCN guideline-based staging categories, domain experts retrospectively mapped the descriptions into one of four defined stages: \textit{Resectable}, \textit{Borderline Resectable}, \textit{Locally Advanced}, or \textit{Metastatic}. We conducted label assignment independently of model development or evaluation procedures.

Three board-certified specialists (from gastroenterology, surgery, and radiology) independently reviewed the report bodies and assigned clinical staging labels to benchmark system performance. They did not view the original conclusions and received no additional guidance or support. We performed no inter-annotator discussion or consensus; each specialist made independent decisions. When a report lacked sufficient information for confident staging, annotators were allowed to assign an ``indeterminate'' label.

We excluded cases labeled as indeterminate in the ground truth from the accuracy calculation for evaluation. Specifically, we removed three free-text reports and two structured reports. We included all remaining cases in the final evaluation.

Due to institutional policies and patient privacy regulations, we are unable to publicly release the dataset used in this study.

\subsection{Experimental Conditions}

We conducted experiments using a hybrid system composed of a cloud LLM (\texttt{gemini-2.5-pro-preview-03-25}) and a local LLM (\texttt{gemma3:27b-it-qat\footnote{\url{https://ollama.com/library/gemma3:27b-it-qat}}}). The cloud model was run with an inference temperature of 0.8 to encourage diverse and creative prompt generation. The local model was executed on an internal GPU server within the hospital network using the Ollama \footnote{\url{https://github.com/ollama/ollama}} inference framework on an RTX 6000 Ada GPU (48GB VRAM), with \texttt{num\_ctx} set to 32k, an inference temperature of 0.2, and structured output mode enabled to produce consistent, machine-readable JSON results.
Processing the full set of 100 radiology reports with MedOrchestra took approximately one hour in total. This was conducted on a single GPU without parallelization.

We designed three experimental settings to evaluate the system:

\begin{itemize}
    \item \textbf{Local LLM (Base)}: The local language model performed staging based solely on the report text, without access to external references such as the NCCN guideline or specific prior training on this task.

    \item \textbf{Local LLM (with Guideline)}: The same local model received the full NCCN guideline document as additional context during staging.

    \item \textbf{MedOrchestra}: The hybrid system decomposed the staging task into clinical subtasks using the cloud LLM, which analyzed the guideline and user input to generate detailed system prompts for feature extraction (e.g., vascular involvement, distant metastasis). The local LLM then executed these prompts to extract relevant clinical features from each report. The system synthesized the extracted features into a final staging prediction using rule-based logic defined by the cloud LLM based on the NCCN guideline.
\end{itemize}

We ran MedOrchestra five times per case and selected the final prediction via majority voting over the five outputs. All clinical inference was performed in a fully isolated, network-disconnected environment. To ensure data privacy and separation, we manually transferred the cloud-generated prompts to this environment in structured JSON format.

\subsection{Evaluation Protocol}

We evaluated performance as a 4-way classification task using the NCCN-defined staging categories. The model was required to assign exactly one of these labels for each case.

We used accuracy as the primary evaluation metric, measuring the proportion of exact matches between model predictions and ground truth labels. For MedOrchestra, we obtained five predictions per case and selected the final output via majority voting. Following a conservative assumption, we chose the label with the higher clinical stage in cases where a tie occurred.

To validate system performance, we conducted two types of comparisons. First, we compared MedOrchestra against a local LLM with no access to external domain knowledge to assess baseline capability. Second, to evaluate clinical plausibility, we compared MedOrchestra's predictions against those of three individual board-certified specialists.

\section{Results}

\begin{figure*}[t]
    \includegraphics[width=1.0\textwidth]{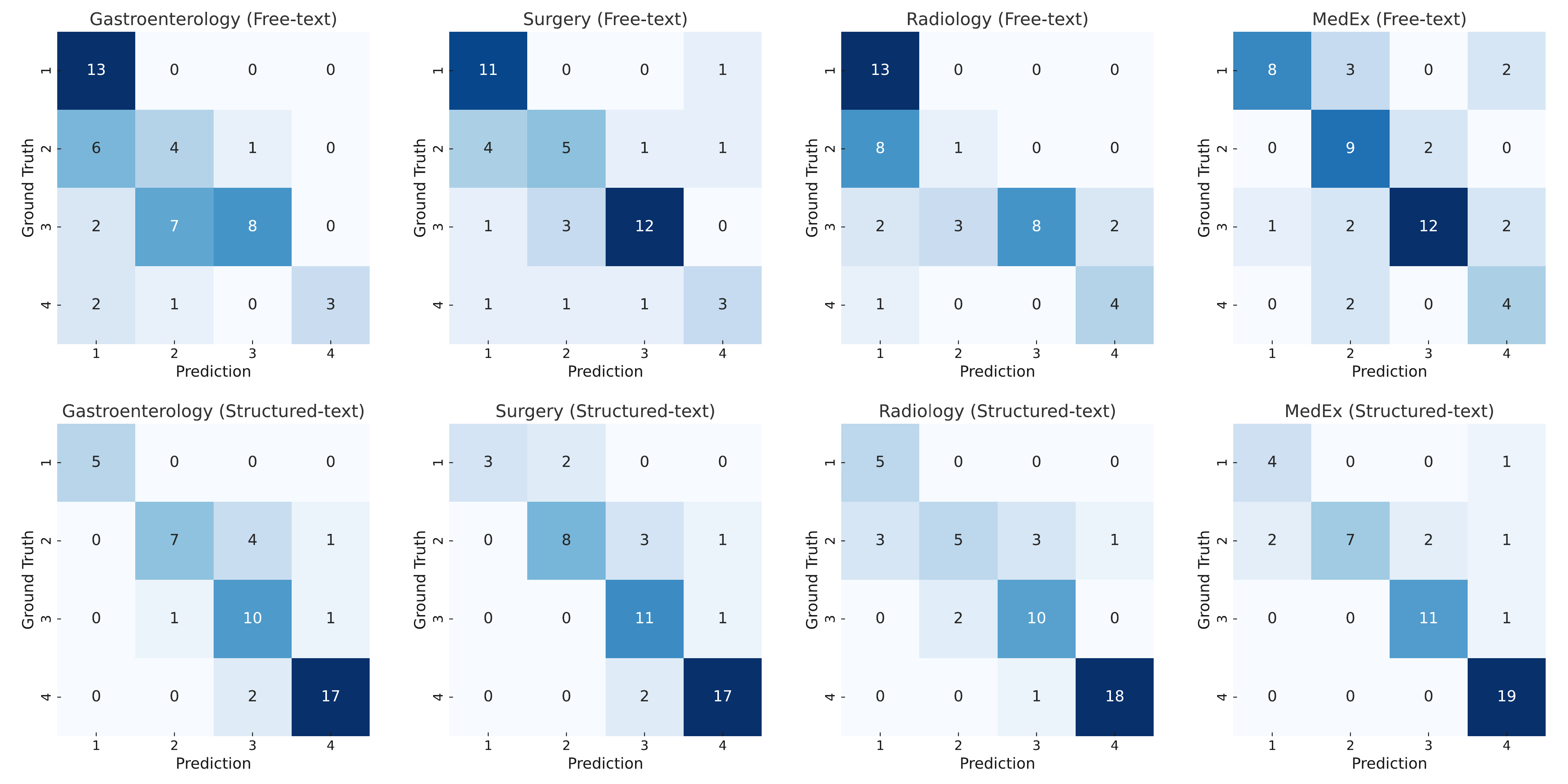} \hfill
    \caption {Confusion matrices for staging predictions (1: Resectable, 2: Borderline Resectable, 3: Locally Advanced, 4: Metastasis) from both free-text (top row) and structured-text (bottom row) radiology reports by three clinical specialists and MedOrchestra.}
    \label{fig:staging_confusion}
\end{figure*}

\begin{figure}[t]
    \includegraphics[width=\linewidth]{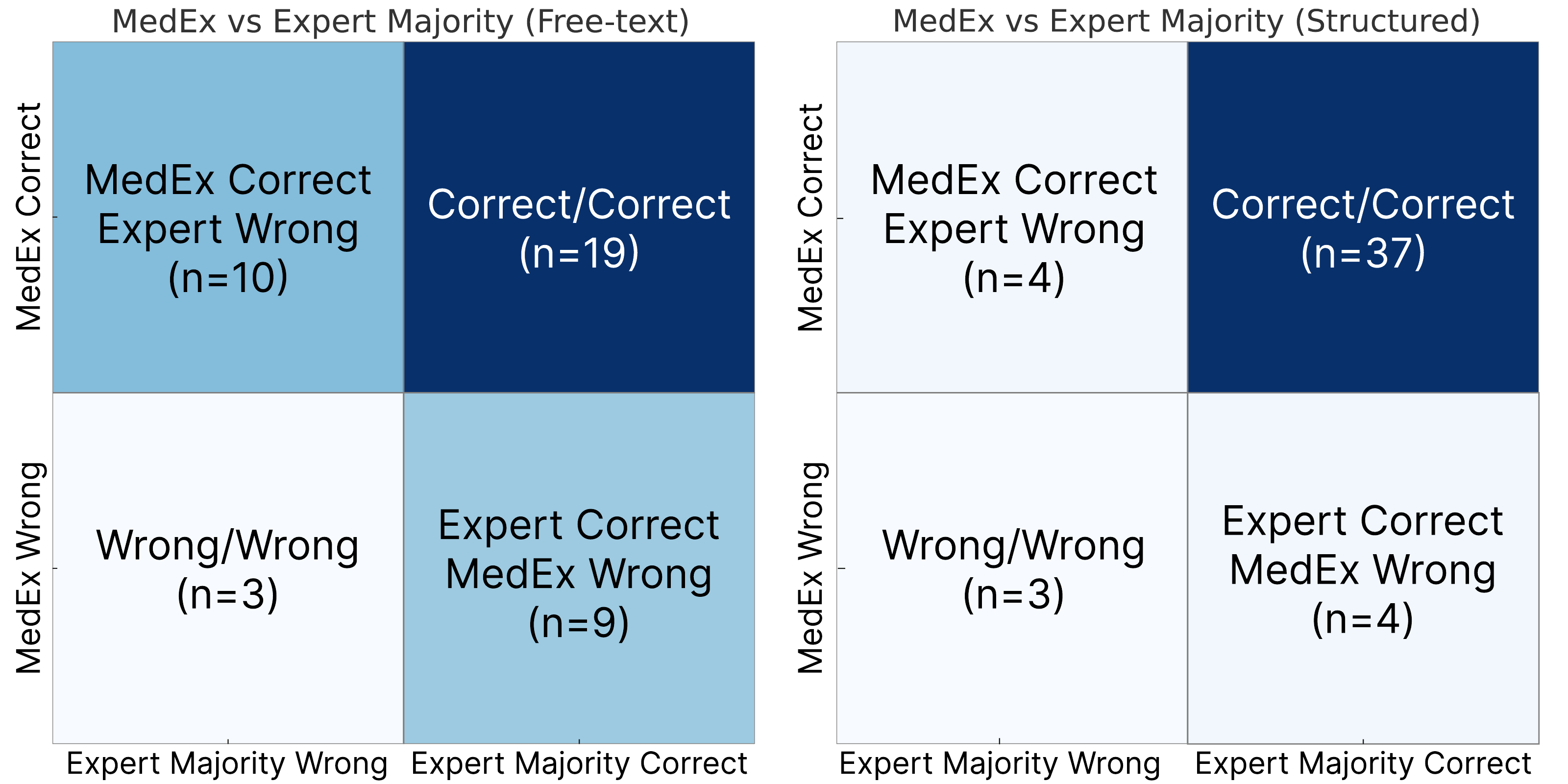} \hfill
    \caption {Comparison of MedOrchestra and expert majority decisions on clinical staging using free-text (left) and structured (right) radiology reports.}
    \label{fig:2x2}
\end{figure}

\begin{table}[ht]
  \centering
  \resizebox{\linewidth}{!}{%
    \begin{tabular}{lcc}
      \hline
      \textbf{Condition} & \textbf{Accuracy (Free-text)} & \textbf{Accuracy (Structured-text)} \\
      \hline
      Local LLM (Base) & 48.94\% & 60.40\% \\
      Local LLM (with Guideline) & 56.59\% & 77.10\% \\
      MedOrchestra (Ours) & \textbf{70.21\%} & \textbf{85.42\%} \\
      Gastroenterologist & 59.57\% & 81.25\% \\
      Radiologist & 55.32\% & 79.17\% \\
      Surgeon & 65.96\% & 81.25\% \\
      \hline
    \end{tabular}
  }
  \caption{Accuracy of each system and expert group on the clinical staging task for both free-text and structured-text radiology reports.}
  \label{tab:accuracy}
\end{table}

\begin{table}[ht]
  \centering
  \resizebox{\linewidth}{!}{%
  \begin{tabular}{lcc}
    \hline
    \textbf{Condition} & \textbf{Kappa (Free-text)} & \textbf{Kappa (Structured-text)} \\
    \hline
    MedOrchestra (Ours) & \textbf{0.596} & \textbf{0.792} \\
    Gastroenterologist & 0.444 & 0.735 \\
    Radiologist & 0.469 & 0.709 \\
    Surgeon & 0.571 & 0.733 \\
    \hline
  \end{tabular}
  }
  \caption{Cohen's Kappa scores indicating agreement with GT clinical staging. Local LLM results are omitted as inter-rater reliability is not applicable.}
  \label{tab:kappa}
\end{table}

This section compares the proposed hybrid system, \textbf{MedOrchestra}, against baseline Local LLM settings and human expert annotations. We assess performance using accuracy (Table~\ref{tab:accuracy}), agreement with GT labels (Cohen’s Kappa; Table~\ref{tab:kappa}), stage-level prediction consistency (Figure~\ref{fig:staging_confusion}), and disagreement analysis between MedOrchestra and expert majority judgments (Figure~\ref{fig:2x2}). We report all results separately for free-text and structured-text inputs.

\subsection{Performance of Local LLMs and the Effect of Clinical Context}

Table~\ref{tab:accuracy} shows the accuracy of two Local LLM baselines: \textit{Local LLM (Base)}, which uses only the input report, and \textit{Local LLM (with guideline)}, which incorporates the complete NCCN guideline as additional context. The Base configuration achieved 48.94\% in the free-text setting, and the guideline-augmented model achieved 56.59\%. In contrast, MedOrchestra achieved 70.21\%, outperforming the two baselines by 21.3 and 13.6 percentage points, respectively. We observed a similar pattern for structured-text inputs. MedOrchestra achieved 85.42\%, outperforming the baselines by 25.0 and 8.3 percentage points.

The Local LLM failed to perform the necessary multi-step reasoning, even with access to the complete guideline. Clinical staging requires coordinated inference over interdependent features such as vascular invasion, organ involvement, and distant metastasis. End-to-end prompting with unstructured context did not support such inference effectively.

MedOrchestra overcomes this limitation by decomposing the task into subtasks. The cloud LLM identifies relevant features and generates structured prompts. The Local LLM extracts the corresponding information, and the system determines the final stage using rule-based logic derived from the NCCN guideline. This pipeline enables more accurate and stable predictions than either baseline.

\subsection{Comparison with Expert Annotations}

Table~\ref{tab:accuracy} shows that MedOrchestra consistently outperformed all expert groups on free-text and structured-text inputs. In the free-text setting, expert accuracies ranged from 55.32\% (Radiologist) to 65.96\% (Surgeon), while MedOrchestra achieved 70.21\%. In the structured-text setting, MedOrchestra again achieved the highest accuracy at 85.42\%.

Table~\ref{tab:kappa} presents the corresponding Cohen’s Kappa scores with GT labels. MedOrchestra achieved the highest agreement in both settings (0.596 for free-text, 0.792 for structured-text), surpassing the best expert performance (0.571 and 0.733). These results show that MedOrchestra achieves higher accuracy and provides more consistent stage assignments relative to the GT.

\subsection{Stage-Level Prediction Consistency}

Figure~\ref{fig:staging_confusion} presents confusion matrices for MedOrchestra and the expert groups. In the free-text setting, experts frequently confused Stage~2 and Stage~3. The Radiologist group often misclassified Stage~2 as Stage~1.

MedOrchestra aligned more closely with GT labels overall, but showed slightly lower accuracy on resectable cases than the experts. Manual review revealed that MedOrchestra tended to interpret speculative expressions (e.g., ``likely,'' ``suspicious for'') as definitive indicators of advanced disease, which led to overstaging. In contrast, experts treated such language as inconclusive and assigned more conservative stage labels.

In the structured-text setting, MedOrchestra correctly predicted all Stage~4 cases (19/19) and showed balanced accuracy across all stages. The confusion matrix exhibited strong diagonal dominance, indicating robust staging consistency.

\subsection{Disagreement Analysis Between MedOrchestra and Expert Majority}

To analyze prediction differences in more detail, we examined cases where MedOrchestra and the expert majority disagreed (Figure~\ref{fig:2x2}).

In the free-text setting, MedOrchestra correctly classified 10 cases that the expert majority misclassified. These cases typically included long, complex reports with staging-relevant details often buried in unrelated content. Our qualitative review of these cases suggests that MedOrchestra’s structured feature extraction strategy helped isolate staging-relevant information more effectively. This advantage likely stems from the cloud LLM’s task decomposition and targeted prompts for the Local LLM, which reduced distraction from unrelated content.

In contrast, the majority of experts correctly classified nine cases that MedOrchestra misclassified. Most of these involved ambiguous or speculative language. MedOrchestra interpreted such phrases as definitive, leading to overstaging. Conversely, experts responded more cautiously to ambiguity and often selected lower stages consistent with the GT.

In the structured-text setting, disagreements decreased substantially. Only four cases in each off-diagonal category showed disagreement, suggesting that structured input helped humans and models interpret staging cues more consistently.

\subsection{Effect of Report Format on Performance}

All systems and annotators improved when given structured-text input, although the size of the improvement varied. MedOrchestra achieved the most significant gain (+15.2 percentage points). Expert gains ranged between 13.6 and 21.7 points. These results indicate that MedOrchestra leverages structured inputs effectively and adapts well to formalized clinical documentation.

\subsection{Summary}

MedOrchestra outperformed both Local LLMs and domain experts across multiple evaluation metrics. The baseline Local LLMs struggled to apply clinical guidelines effectively, which reflects the limitations of end-to-end prompting for complex reasoning. In contrast, MedOrchestra used task decomposition and rule-based inference to extract relevant features and predict cancer stages accurately. While MedOrchestra performed consistently across formats, handling ambiguity in free-text reports remains an open challenge.

\section{Conclusion}

MedOrchestra is a hybrid clinical NLP framework that combines the reasoning capabilities of cloud LLMs with the privacy-preserving execution of local models. Our framework addresses the critical gap between the limited reasoning capacity of local LLMs for complex tasks such as cancer staging and the data governance challenges associated with cloud LLMs. MedOrchestra decomposes high-level clinical decisions into structured subtasks, which are executed locally using prompts generated by the cloud LLM, enabling accurate and interpretable inference under secure deployment settings.

We demonstrated superior performance to both local LLM baselines and clinical expert groups on pancreatic cancer staging. In particular, it showed strong results in free-text settings, where reports tend to be long, unstructured, and contain extraneous information. MedOrchestra was able to reliably extract relevant features and apply guideline-based logic, even in these challenging contexts. While structured inputs yielded higher absolute accuracy, the system’s consistent performance on free-text data underscores its practical utility in real-world clinical documentation.

The proposed framework shows potential for broader application to other guideline-based clinical decision-making tasks. Future work will focus on refining its handling of ambiguous or speculative language, evaluating its applicability in new clinical domains, and exploring integration with multimodal clinical data. MedOrchestra offers a practical and extensible architecture for deploying LLMs in clinical environments with accuracy, interpretability, and privacy in balance.

\section*{Limitations}
While this study demonstrates the potential of a hybrid LLM framework for clinical data processing, several important limitations warrant consideration:

\textbf{Limited Scope and Generalizability.} We evaluated the framework on 100 radiology reports from pancreatic cancer patients at a single institution, focusing specifically on staging tasks by well-defined NCCN guidelines. The study covers a single disease type and clinical context, which limits its breadth. The framework works best for clinical tasks with explicit, structured guidelines and may struggle in domains where guidelines remain ambiguous or nonexistent. We have yet to verify its generalizability across other diseases, institutions, and data formats.

\textbf{Local LLMs Performance Constraints.} Although we decomposed the overall task into smaller subtasks, local LLMs still show performance gaps compared to cloud LLMs when handling complex narratives. While feature extraction helps mitigate the issue, some clinical guidelines require higher-level reasoning, such as understanding temporal progression, inferential logic, which simple decomposition cannot effectively capture.

\textbf{Ground Truth Ambiguity and Input Quality Issues.} Defining a consistent Ground Truth (GT) for clinical staging is fundamentally challenging, as some imaging cases remain ambiguous even among specialists. Different clinicians may interpret the same image differently, especially when clear diagnostic evidence is lacking. Furthermore, approximately 20 to 30 percent of the free-text radiology reports in our dataset did not contain sufficient supporting detail outside the conclusion section. While the conclusion often stated the stage enough to assign a GT, the earlier sections of the report, such as findings and impressions, often lacked the necessary details. In cases where the report lacked sufficient information outside the conclusion, determining the stage became difficult, which limited the reliability of GT construction and model evaluation.

\textbf{Operational Infrastructure Challenges.} The hybrid framework depends on interaction between cloud and local LLMs, but clinical systems often restrict external network access due to security policies. Because of these restrictions, users cannot run cloud-based tasks directly within the clinical environment. Instead, they must perform tasks like decomposition and instruction generation externally and manually transfer the system prompts into the internal system (local LLMs). This segmented workflow increases operational burden and limits seamless integration.

\textbf{Insufficient Validation of Multi-Round Inference.} To improve consistency in local LLM outputs, we applied repeated inference with majority voting and low temperature settings. However, we did not perform a systematic validation to determine the optimal number of repetitions or to assess output consistency across runs. Future work should introduce clear metrics to evaluate the effectiveness and reliability of multi-round inference strategies.

\section*{Ethics Statement}
This study prioritizes patient privacy by ensuring that no sensitive clinical data is transmitted to external servers. All real data processing is performed in a fully isolated local environment, while the cloud-based LLM is used only for meta-level operations such as task decomposition and prompt generation, without access to actual patient records.

Nonetheless, several potential risks remain. First, the system may overinterpret ambiguous or speculative language in free-text reports, which can lead to overstaging. Second, the evaluation is limited to a single institution and disease type (pancreatic cancer), limiting generalizability and introducing potential bias. Third, practical deployment in clinical settings requires manual prompt transfer due to institutional network restrictions, increasing operational burden.

While the system is designed to support expert decision-making, there remains a risk that it may be used to make clinical decisions autonomously in practice. To mitigate this risk, future work should investigate mechanisms to explicitly require and structurally integrate expert oversight throughout the framework, ensuring safe and responsible deployment in real-world clinical environments.


\end{document}